\documentclass{IEEEcsmag}

\usepackage[colorlinks,urlcolor=blue,linkcolor=blue,citecolor=blue]{hyperref}
\usepackage{cite}
\usepackage{amsmath,amssymb,amsfonts}
\usepackage{graphicx}
\usepackage{textcomp}
\usepackage{xcolor}
\usepackage{algorithm,algorithmic}

\usepackage{array}
\usepackage{booktabs}
\usepackage{makecell}
\usepackage{subfigure} 
\usepackage{cleveref}
\usepackage{stfloats}
\expandafter\def\expandafter\UrlBreaks\expandafter{\UrlBreaks\do\/\do\*\do\-\do\~\do\'\do\"\do\-}
\usepackage{upmath,color}

\begin{document}

\sptitle{FEATURE ARTICLE: DEEP REINFORCEMENT LEARNING}

\title{Model predictive control--based \\
value estimation for \\
efficient reinforcement learning}
\author{Qizhen Wu}
\affil{Beihang University, Beijing, 100191, China}
\author{Kexin Liu}
\affil{Beihang University, Beijing, 100191, China}
\author{Lei Chen}
\affil{Beijing Institute of Technology, Beijing, 100081, China}
\markboth{THEME/FEATURE/DEPARTMENT}{THEME/FEATURE/DEPARTMENT}

\begin{abstract}\looseness-1Reinforcement learning suffers from limitations in real practices primarily due to the number of required interactions with virtual environments. It results in a challenging problem because we are implausible to obtain a local optimal strategy with only a few attempts for many learning methods. Hereby, we design an improved reinforcement learning method based on model predictive control that models the environment through a data--driven approach. Based on the learned environment model, it performs multi--step prediction to estimate the value function and optimize the policy. The method demonstrates higher learning efficiency, faster convergent speed of strategies tending to the local optimal value, and less sample capacity space required by experience replay buffers. Experimental results, both in classic databases and in a dynamic obstacle avoidance scenario for an unmanned aerial vehicle, validate the proposed approaches.
\end{abstract}

\maketitle

\chapteri{T}he increasing applications of reinforcement learning (RL) in fields such as game playing$^{1}$, natural language processing$^{2}$, and robotics$^{3}$ gain much attention due to its advancements in artificial intelligence. 
However, the low sample utilization of RL is challenging in applications, since agents have limited interactions with the environment.
It means that $n$--step temporal difference ($n$--TD)$^{4}$, as a typical representation of model--free RL aiming to enhance the learning efficiency with mass environmental interactions, is not adaptive to improve the sample utilization$^{5}$.

To overcome this difficulty, model--based RL (MBRL) generates virtual data$^{6}$ and combines with model predictive control (MPC) to achieve decision making with fewer attempts and computation$^{7}$. The improved RL methods based on MPC apply multi--step prediction for the interactions with virtual environments, which generates additional data to improve RL efficiency. Model--based policy optimization (MBPO) uses the probabilistic ensemble approach to approximate the environment model$^{8}$. It implements stepwise inference of the environment model based on MPC and adopts the model data obtained from the inference for policy training. Furthermore, masked model--based actor--critic implements a masking mechanism based on the model’s uncertainty to determine the availability of its prediction$^{9}$. This approach takes complete account of the complexity of the model but reduces computational speed due to the large number of neural networks.

In addition to generating more samples, MPC--based RL can better utilize virtual environments by improving single--step value estimation of RL$^{10}$. It leverages the iterative data generated from virtual environments to construct an intensified value, which is suffer deeply from inaccurate models$^{11}$.
Therefore, TD--MPC employs a neural network that simultaneously approximates the Q--function and the environment model$^{12}$.
A novel MBRL approach adopts filtered probabilistic MPC to model the unobservable disturbances of the environment$^{13}$. 
Most existing ways to improve the estimation of the value function are based on short--term rewards but do not well balance the impact of long--term rewards.

In this study, we present a novel MPC--based RL method that focuses on both improving the value estimation and modeling the environment to enhance the learning efficiency and sample utilization of intelligent agents. Our method performs multi--step prediction to estimate the value function and optimize the policy. It uses a deterministic model--based approach to approximate the environment and applies a rolling optimization approach$^{14}$ to maximize the cumulative return for each prediction interval. 
In applications, we conduct experimental comparisons with baselines in classic simulation environments
and a practical RL problem of unmanned aerial vehicle (UAV) dynamic obstacle avoidance.
In comparison, our method leads the strategy to quickly converge to the local optimal value based on fewer interaction data. 
We verify that the state transition and reward function models we trained approximate the environment model well, especially in the case of low--dimensional state and action spaces. Furthermore, a poorly fitted environment model may cause the policy to converge to a sub--optimal rather than a global optimal value more quickly in the high--dimensional problem.

We organize the rest of the paper as follows. The section "Main Results" presents our proposed method. The section "Experimental Results" describes the experiments of our method. Finally, in the section "Conclusions", we present our conclusions.
\vspace*{-5pt}

\section{MAIN RESULTS}
\label{sec:sample2}
\subsection{Presentation of Basic Algorithms}

\looseness-1Let $\mathbb{R}$ denote the set of real numbers. $\mathbb{E}$ denotes the mathematical expectation.

RL can be modeled by a markov decision process (MDP). It is represented by a five--tuple $\left( S,A,P,R,\gamma \right)$, where $S\subseteq \mathbb{R}^{m_s}$ is the environment state space and $A\subseteq \mathbb{R}^{m_a}$ is the action space. $m_s$ and $m_a$ denote the state and action space dimensions, respectively. $P:S\times A\rightarrow \mathbb{R}^{m_s}$ denotes the state transition function, $R:S\times A\rightarrow \mathbb{R}$ denotes the reward function, and $\gamma$ denotes the discount factor. The purpose of RL is to optimize the policy $\pi :S\rightarrow A$ such that the cumulative reward $\mathbb{E}_{\varGamma \sim\pi }\left[ \sum_{t=1}^{\infty}{\gamma ^tr_t} \right] ,r_t \sim R\left( s_t,a_t \right)$ is maximized. $\varGamma =\left( s_0,a_0,s_1,a_1,... \right)$ is the trajectory of the agent interacting with the environment under the policy, where $a_t\sim\pi \left( s_t \right)$ and $s_{t+1}\sim P\left( s_t,a_t \right)$ denote the selected action and reaching state at each decision step $t$, respectively.

In the temporal difference method$^5$, the value--based RL approximates the cumulative rewards $\sum_{t=1}^{\infty}{\gamma ^tr_t}$ by $R\left( s,a \right) +\gamma \max Q^*\left( s',a' \right)$. It estimates the optimal state--action value function $Q^*:S\times A\rightarrow \mathbb{R}$ through $Q$ table or a parameterized neural network $Q_{\omega}\left( s,a \right) \approx Q^*\left( s,a \right) =\mathbb{E}\left[ R\left( s,a \right) +\gamma \max Q^*\left( s',a' \right) \right] ,\forall s\in S$, where $s',a'$ are the state and action at the following step, respectively. $\omega$ is the weighting factor of the neural network. For $\gamma \approx 1$, $Q^*$ estimates the discounted returns of the local optimal strategy over an infinite range. For value--based RL, the value estimation of $Q\left( s,a \right)$ can be described as the following form:
\begin{align}
        Q\left( s,a \right) \gets~ &Q\left( s,a \right) \nonumber \\
        &+\alpha \left[ r +\gamma Q\left( s',a' \right) -Q\left( s,a \right) \right],
        \label{equ1}
\end{align}
where $\alpha$ is the learning rate of policy training.
For value--based deep--RL, $Q^*$ can be approximated by an iteratively fitting $Q_{\omega}$ and the loss function is described as:
\begin{equation}
    \begin{array}{c}
        L_{\omega}=\mathbb{E}_{\left( s,a \right) \sim \mathfrak{B}}\lVert Q_{\omega}\left( s,a \right) -y \rVert ^2,
    \end{array}\label{equ2}
\end{equation}
where $y=R\left( s,a \right) +\gamma \max Q_{\omega ^-}\left( s',a' \right)$ is the $Q-target$. The subscript $\omega^{-}$ is a slow moving online average. $\omega^{-}$ is updated with the rule $\omega _{k+1}^{-}\gets \left( 1-\zeta \right) \omega _{k}^{-}+\zeta \omega _k$ at each iteration using a constant factor $\zeta \in \left[ 0,1 \right)$. $\mathfrak{B}$ is a replay buffer that iteratively grows as data are updated. 

In the actor--critic algorithm, $\pi$ is usually a policy parameterized by a neural network $\pi_{\beta}$. It learns the approximation $\pi_{\beta}\left( s  \right) \approx arg\max_a\mathbb{E}\left[ Q_{\omega}\left( s,a \right) \right] ,\forall s\in S$, which is the local optimal policy. The loss function is:
\begin{equation}
    \begin{array}{c}
        L_{\beta}=-\mathbb{E}_{\left( s,a \right) \sim \mathfrak{B}}\left[Q_{\omega}\left( s,a \right)\right].
    \end{array}\label{equ23}
\end{equation}

In control theory, we formulate the state space equation as follows:
\begin{align}
        \dot{s} = As + Ba,
        \label{equ4}
\end{align}
where $s$ and $a$ are the state and action vectors, respectively. MPC obtains a local solution to the trajectory optimization problem at each step $t$ by estimating the local optimal action $a_{t:t+N}$ on a finite horizon $N$ and executing the first action:
\begin{figure*}[htbp]
    \centering
    \includegraphics[height=0.25\textwidth]{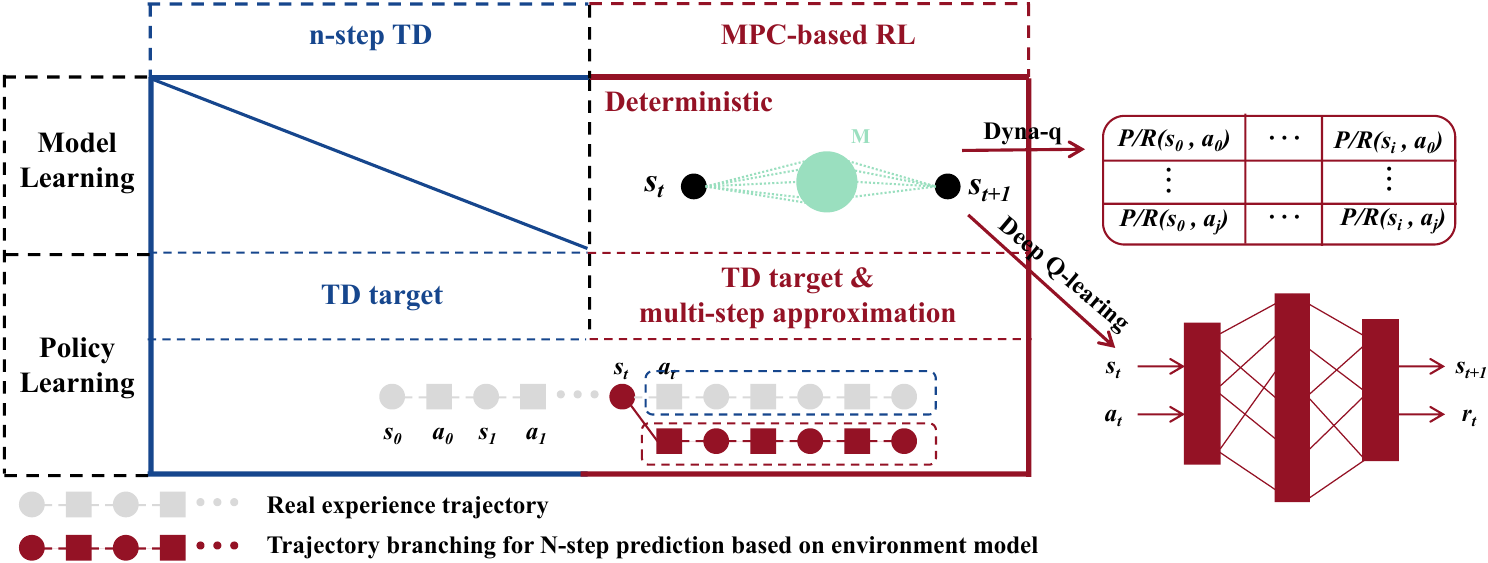}
    \caption{Differences between $n$--TD and our method.}
    \label{fig1}
\end{figure*} 
\begin{figure*}[ht]
    \centering
    \includegraphics[height=0.25\textwidth]{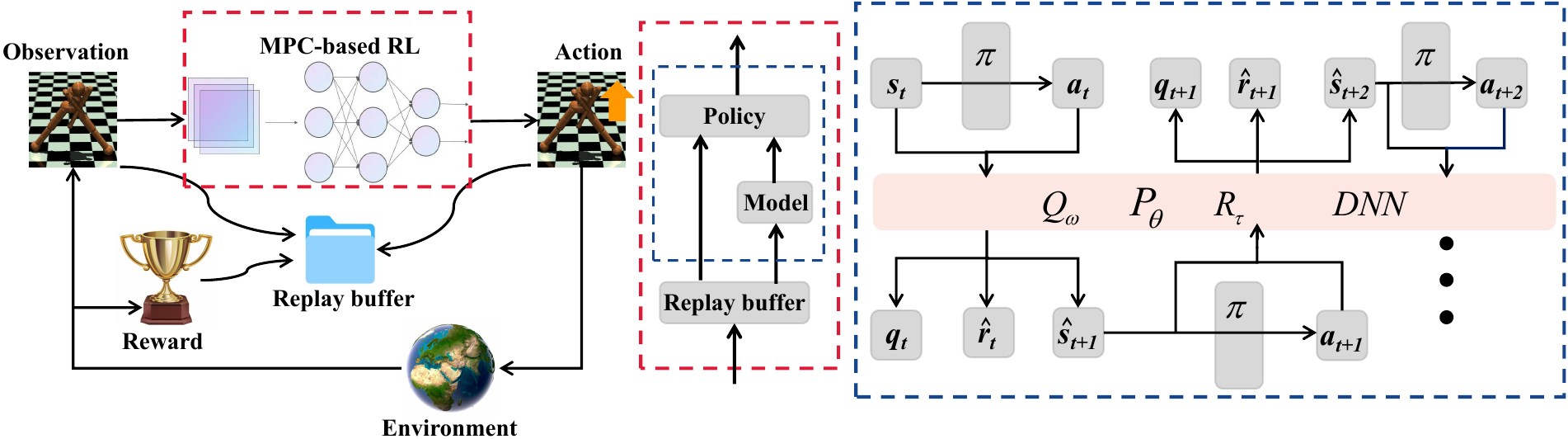}
    \caption{Framework of MPC--based RL.}
    \label{fig2}
\end{figure*} 
\begin{align}
        \pi_{1}^{MPC}\left( s_t \right) =~&\underset{a_{t:t+N}}{\arg\min} ~\mathbb{E} \bigg[\sum_{i=t}^{t+N-1}{\left(s_{i}^{T}G~s_i +a_{i}^{T}E~a_i\right)} \bigg], 
        \label{equ5} \\ 
        \pi_{2}^{MPC}\left( s_t \right) =~&\underset{a_{t:t+N}}{\arg\min} ~\mathbb{E} \bigg[\sum_{i=t}^{t+N-1}{\left(s_{i}^{T}G~s_i +a_{i}^{T}E~a_i\right)} \nonumber \\
        & +s_{N}^{T}P~s_N \bigg],
        \label{equ6}
\end{align}
where $G$, $E$, and $P$ are error weighting, control weighting, and terminal error weighting matrix, respectively. (\ref{equ5}) is the basic MPC and (\ref{equ6}) is the basic MPC with added terminal cost$^{14}$.

\subsection{The Improved MPC--based RL Algorithm}
To enhance the learning efficiency and reduce the interaction data required, we propose an online training method focusing on improving the value estimation and the modeling of the environment. Our method performs multi--step prediction to estimate the value function and optimize the policy. It learns local optimal strategies by modeling the environment's state transition and reward functions using a data--driven approach. Moreover, it employs a rolling optimization approach to maximize the cumulative return for each prediction interval.

MPC--based RL replaces process and terminal cost with rewards and terminal value functions. By transforming the problem from least cost to most benefit in (\ref{equ5}) and (\ref{equ6}), it obtains the following forms:
\begin{align}
        \pi_{1}^{MPC-RL}\left( s_t \right) =~&\underset{a_{t:t+N}}{\arg\max} ~\mathbb{E} \bigg[ \sum_{k=t}^{t+N-1}{\gamma ^kR\left( s_k,a_k \right)} \bigg],
        \label{equ7} \\ 
        \pi_{2}^{MPC-RL}\left( s_t \right) =~&\underset{a_{t:t+N}}{\arg\max} ~\mathbb{E} \bigg[ \sum_{k=t}^{t+N-1}{\gamma ^kR\left( s_k,a_k \right)} \nonumber \\
        &+\gamma^{t+N}\max Q_{\theta ^-}\left( s_{t+N},a_{t+N} \right) \bigg].
        \label{equ8}
\end{align}

As shown in Figure \ref{fig1}, the main difference between $n$--TD and our method is that, in (\ref{equ7}) and (\ref{equ8}), the states and rewards at moments $t+1$ to $t+N$ are obtained from real experience trajectory. In contrast, the states and rewards of MPC--based RL are obtained from trajectory branching for $N$--step prediction based on environment model. 
MPC--based RL not only improves the TD target, but also employs a multi--step approximation approach to estimate the value function.

The goal of strategy in MPC is to predict a sequence in $N$ steps of actions such that the short--term cumulative rewards are maximized in such interval. The goal of policy in RL is to maximize the long--term cumulative rewards for all future moments $\mathbb{E}_{\varGamma \sim\pi }\left[ \sum_{k=1}^{\infty}{\gamma ^kr_k} \right] ,r_k \sim R\left(  s_k,a_k \right)$. We replace the reward $R\left( s_k,a_k \right)$ with the cumulative rewards $\sum_{i=k}^{\infty}{\gamma ^ir_i}$ in (\ref{equ7}) that maximizes both short--term and long--term rewards, and the strategy can be described in the following form:
\begin{equation}
    \begin{array}{c}
        \pi ^{MPC-RL}\left( s_t \right) =\underset{a_{t:t+N}}{\arg\max} ~\mathbb{E}\left[ \sum\limits_{k=t}^{t+N}{\gamma ^k\left( \sum\limits_{i=k}^{\infty}{\gamma ^ir_i} \right)} \right].
    \end{array}\label{equ9}
\end{equation}
In (\ref{equ9}), we perform predictive control based on a single moment $t$ to produce a local optimal solution in the present moment. The rolling optimization is then performed to generate multiple local optimal solutions and calculate the optimal value. The use of cumulative rewards compensates for the lack of the terminal reward. We can improve sample utilization and training efficiency by accumulating the cumulative rewards in the next $N$ steps.

The interaction results in the next $N$ moments are required in (\ref{equ9}), but the sample data stored in the experience replay buffer $\mathfrak{B}$ only have the interaction tuple for a single moment $t$. The models are unknown for most interaction environments, and these sample data cannot be inferred from a priori knowledge. Therefore, we need to build neural networks with states transition function $P$ and reward function $R$ to predict the interaction outcome of the next $N$ moments from the input states and actions. $P$ and $R$ can be modeled as separate neural networks ($P_{\theta},R_{\tau}$) or combined into one network ($PR_{\psi}$). The subscripts $\theta,\tau$, and $\psi$ are the weighting factors of neural networks. We use a deterministic model--based approach instead of the probabilistic ensemble$^{15}$. It reduces the number of neural networks and improve the computational speed of the algorithm, albeit at the cost of partially ignoring model uncertainty. 

In our framework shown in Figure \ref{fig2}, the samples drawn from the experience replay buffer are used to update the environment model and the policy. Meanwhile, the environment model has a facilitating effect on the update of the policy model. Our approach performs multi--step prediction based on one sample data, resulting in faster convergence of the strategy. Similar to RL, we approximate the cumulative rewards $\sum_{t=1}^{\infty}{\gamma ^tr_t}$ by $R\left( s,a \right) +\gamma \max Q^*\left( s',a' \right)$. Instead of letting $Q\left( s_k,a_k \right)$ be close to $y_k$, our method performs a multi--step approximation approach: 
\begin{gather}
    \left\{ \begin{array}{c}
	\displaystyle Q\left( s_k,a_k \right) \rightarrow y_k\\
	\displaystyle Q\left( \hat{s}_{k+1},a_{k+1} \right) \rightarrow y_{k+1}\\
	\vdots\\
	\displaystyle Q\left( \hat{s}_{k+n},a_{k+n} \right) \rightarrow y_{k+n}\\
\end{array} \right. ,
        \label{equ14}  
\end{gather}
where $\hat{s},\hat{r}$ indicate that the state and reward are predicted based on the model we learned.

In RL, we define the value estimation as (\ref{equ1}), which is a single--step policy updating. Based on the idea that MPC performs multi--step prediction in the forecast interval, we can improve the value estimation of $Q\left( s,a \right)$ by:

\begin{algorithm}[htb]
    \renewcommand{\algorithmicrequire}{\textbf{Input:}}
	\renewcommand{\algorithmicensure}{\textbf{Output:}}
	\caption{MPC--based RL}
    \label{alg2}
    \begin{algorithmic}[1] 
        \REQUIRE  Enviorment($Env$); Action space($Act$); \\
        Discount factor($\gamma$); Learning rate($\alpha$); \\
        Training episode($E$);Training step($T$);\\
        Prediction step($N$);
	    \ENSURE Policy--network($\pi \left( s \right)$) 
        
        \STATE Initialize networks $\omega,\theta,\tau,\psi$ and the replay buffer $\mathfrak{B}$;
        \FORALL {$e\ =\ 1\rightarrow E$}
            \STATE Get the initial state $s_0$;
            \FORALL {$t\ =\ 1\rightarrow T$}
                \STATE  $a_{t} \gets \pi \left( s \right)$, $\epsilon$--greedy;
                \COMMENT{on-policy action}
                \STATE $r_t,s_{t+1} \gets Env$;
                \COMMENT{transition in environment}
                \STATE $\mathfrak{B} \gets (s_t,a_t,r_t,s_{t+1})$; 
                \COMMENT{replay buffer update}
                \STATE Randomly selected $B$ samples $(s_k,a_k,r_k,s_{k+1})$ from $\mathfrak{B}$;
                \STATE $\hat{s}_{k+1},\hat{r}_{k} \gets$ 
                \COMMENT{prediction in model}\\
                $P_\theta\left( s_{k},a_{k} \right),R_\tau\left( s_{k},a_{k} \right)$\\
                or $PR_\psi\left( s_{k},a_{k} \right)$;
                \IF{$L_{\theta}, L_{\tau}< \epsilon_m$ or $L_{\psi}< \epsilon_m$}
                    \FORALL {$n\ =\ 1\rightarrow N$}
                        \STATE $a_{k+n} \gets \pi \left( s_{k+n} \right)$;
                        \STATE $\hat{s}_{k+n+1},\hat{r}_{k+n} \gets$ 
                        \COMMENT{multi-step prediction}\\
                        $P_\theta\left( s_{k+n},a_{k+n} \right),R_\tau\left( s_{k+n},a_{k+n} \right)$\\
                        or $PR_\psi\left( s_{k+n},a_{k+n} \right)$;
                    \ENDFOR
                \ELSE
                \STATE $a_{k+1} \gets \pi \left( s_{k+1} \right)$;
                \ENDIF
                \STATE Calculate $L_{\omega}$ through (\ref{equ11}) and (\ref{equ12});
                \STATE Calculate $L_{\theta}, L_{\tau}$ through (\ref{equ18});
                \STATE Refresh networks $\omega,\theta,\tau$;
                \COMMENT{gradient-descent}
                \STATE Refresh the strategy $\pi \left( s \right)$;
            \ENDFOR
        \ENDFOR
    \end{algorithmic}
\end{algorithm}
\begin{align}
    &Q\left( s_{k+n},a_{k+n} \right) \gets Q\left( s_{k+n},a_{k+n} \right) \nonumber \\
    &+\alpha \left[ \hat{r}_{k+n}+\gamma Q\left( \hat{s}_{k+n+1},a_{k+n+1} \right) -Q\left( s_{k+n},a_{k+n} \right) \right],
        \label{equ10}
\end{align} 
where $n=0,1,2,...,N-1$ is the current prediction step. We improve the loss function in (\ref{equ2}) as follows:
\begin{equation}
    \begin{array}{c}
        L_{\omega}=\mathbb{E}_{\left( s_k,a_k \right) \sim \mathfrak{B}}\sum\limits_{n=0}^{N-1}{\gamma^n\lVert Q_{\omega}\left( s_{k+n},a_{k+n} \right) -y_{k+n} \rVert ^2}.
    \end{array}\label{equ11}
\end{equation}
According to (\ref{equ8}), we modify the $Q-target$ to the following form:
\begin{align}
        \displaystyle y_{k+n} = ~& \sum\nolimits_{i=n}^{N-1}\gamma ^{i-n}~\hat{r}_{k+i} \nonumber\\ 
        \displaystyle & +  \gamma ^{N-n} \max Q_{\omega ^-}\left( s_{k+N},a_{k+N} \right).
        \label{equ12}
\end{align} 

MPC--based RL can lead the policy to converge to the local optimal value faster. However, the learning speed and accuracy of the environment model can limit the speed and optimality with which the policy finds the local optimal value$^{4}$.

Meanwhile, we need to update the neural networks of the environment model in real--time. The loss functions of $P_{\theta}$ and $R_{\tau}$ are:
\begin{align}
        L_{\theta}&=~\mathbb{E}_{\left( s_k,a_k \right) \sim \mathfrak{B}}\lVert \left( \hat{s}_{k+1}-s_{k+1} \right) \rVert ^2, \nonumber \\ 
        L_{\tau}&=~\mathbb{E}_{\left( s_k,a_k \right) \sim \mathfrak{B}}\lVert \left( \hat{r}_{k}-r_{k} \right) \rVert ^2.
        \label{equ17}
\end{align}
The loss function of $PR_{\psi}$ is:
\begin{align}
        L_{\theta}=\mathbb{E}_{\left( s_k,a_k \right) \sim \mathfrak{B}}\left[\lVert \left( \hat{s}_{k+1}-s_{k+1} \right) \rVert ^2 + \lambda \lVert \left( \hat{r}_{k}-r_{k} \right) \rVert ^2 \right],
        \label{equ71}
\end{align}
where $\lambda > 0$ is a hyper--parameter, which defines the relative weighting of states and rewards.
(\ref{equ17}) and (\ref{equ71}) indicate the degree of deviation of the constructed environment model from the real environment. By minimizing the values of these functions, the environment model can increasingly approximate the real environment. 

We show that the environment model we learned approximates the real environment well in low--dimensional state and action spaces in simulations. However, in high--dimensional environments, the loss functions of environment models may not converge to $0$ and fail to reflect the complete situation of the real environment. This model error can accumulate and lead to a sub--optimal policy. To avoid the impact of inaccurate environment models on the algorithm, we enable environment models for prediction only when their loss functions are less than the certain bounds $\epsilon_m$. The flowchart of the MPC--based RL approach is shown in Algorithm \ref{alg2}.
\begin{table*}[htb]
  \centering
  \caption{The setting of algorithm parameters}
  \label{tab1}
  \begin{tabular*}{35pc}{@{}p{40pt}<{\raggedright}p{35pt}<{\raggedright}p{35pt}<{\raggedright}p{35pt}<{\raggedright}p{35pt}<{\raggedright}p{35pt}<{\raggedright}p{35pt}<{\raggedright}p{35pt}<{\raggedright}p{35pt}<{\raggedright}@{}}
  \toprule
  $Env$ 
  & $\alpha_\omega$ \par ($\times 10^{-2}$)
  & $\alpha_\beta$ \par ($\times 10^{-4}$)
  & $\gamma$ 
  & $\epsilon$ \par ($\times 10^{-2}$)
  & $\mathfrak{B}$ \par ($\times 10^{4}$)
  & $\alpha_\theta$ \par ($\times 10^{-3}$)
  & $\alpha_\tau$ \par ($\times 10^{-3}$)
  & $\alpha_\psi$ \par ($\times 10^{-3}$)
   \\
  \colrule
  CW & 10 & \empty & 0.9 & 1 & \empty & \empty & \empty & \empty\\[3pt]
  CP & 0.2 & \empty & 0.98 & 1 & 1 & 2 & 2 & 2\\[5pt]
  PD & 0.3 & 3 & 0.98 & \empty & 1 & 3 & 3 & 3\\[5pt]
  HO & 0.1 & 1 & 0.99 & \empty & 100 & 1 & 1 & \empty\\[5pt]
  UAV path \par planning & 0.1 & 10 & 0.99 & \empty & 100 & 1 & 1 & 1\\
  \botrule
  \end{tabular*}\vspace*{8pt}
\end{table*}

To account for the monotonic improvement of model--based methods over model--free methods, we can construct a bound of the following form$^{8}$:
\begin{align}
    \eta \left[ \pi \right] \ge ~& \eta ^{branch}\left[ \pi \right] -C, \nonumber \\
    C=~&2r_{\max}\bigg[ \frac{\gamma ^{k+1}\epsilon _{\pi}}{\left( 1-\gamma \right) ^2} \nonumber \\ 
    &+\frac{\left( \gamma ^k+2 \right) \epsilon _{\pi}+N\left( \epsilon _m+2\epsilon _{\pi} \right)}{\left( 1-\gamma \right)} \bigg],
        \label{equ15}  
\end{align}
where $\eta \left[ \pi \right]$ and $ \eta ^{branch}\left[ \pi \right]$ denote the returns of the policy in the MDP and under our model, respectively. $\epsilon _m$ denotes the generalization error due to sampling, and $\epsilon _{\pi}$ denotes the distribution shift due to the updated policy encountering states not seen during model training. We can guarantee that it also improves in the MDP if the strategy enhances its return under the model by at least $C$. MPC--based RL updates the environment model at each decision step $t$. Therefore, the environment model is more accurate and the strategy difference between iterations is more significant. In this case, the generalization error $\epsilon _m$ is much smaller than distribution shift $\epsilon _{\pi}$. The optimal prediction step $N^*$ can be expressed in the following form:
\begin{align}
    N^*=~&\underset{n}{\arg\min} ~\bigg\{ \frac{\gamma ^{k+1}\epsilon _{\pi}}{\left( 1-\gamma \right) ^2} \nonumber \\
    &+\frac{\left( \gamma ^k+2N+2 \right) \epsilon _{\pi}}{\left( 1-\gamma \right)} \bigg\} >0.
        \label{equ16}  
\end{align}

\section{EXPERIMENTAL RESULTS}
\label{sec:sample3}
\begin{figure*}[h]
\centering
\subfigure[]{
\includegraphics[width=0.3\textwidth]{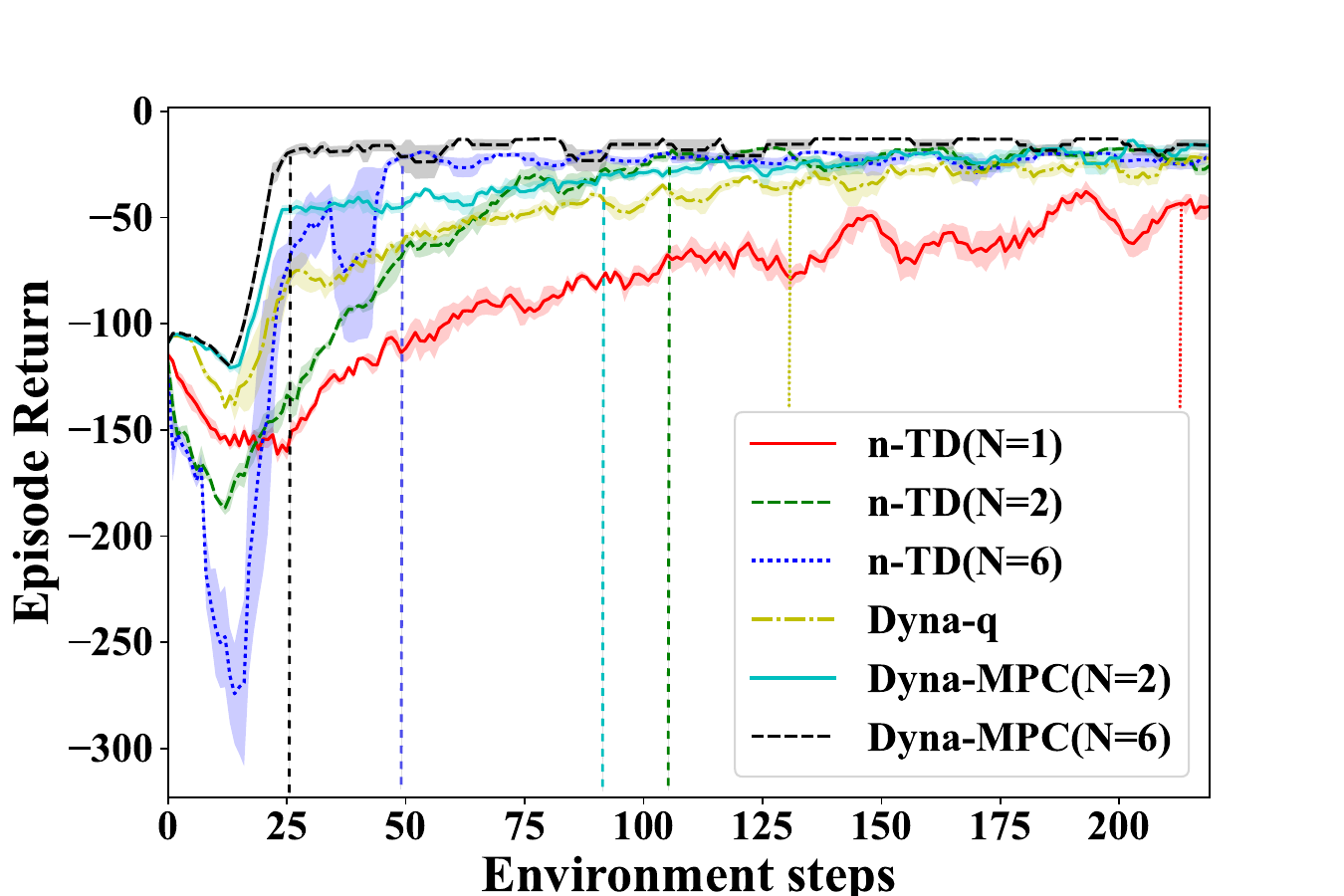} 
}
\subfigure[]{
\includegraphics[width=0.3\textwidth]{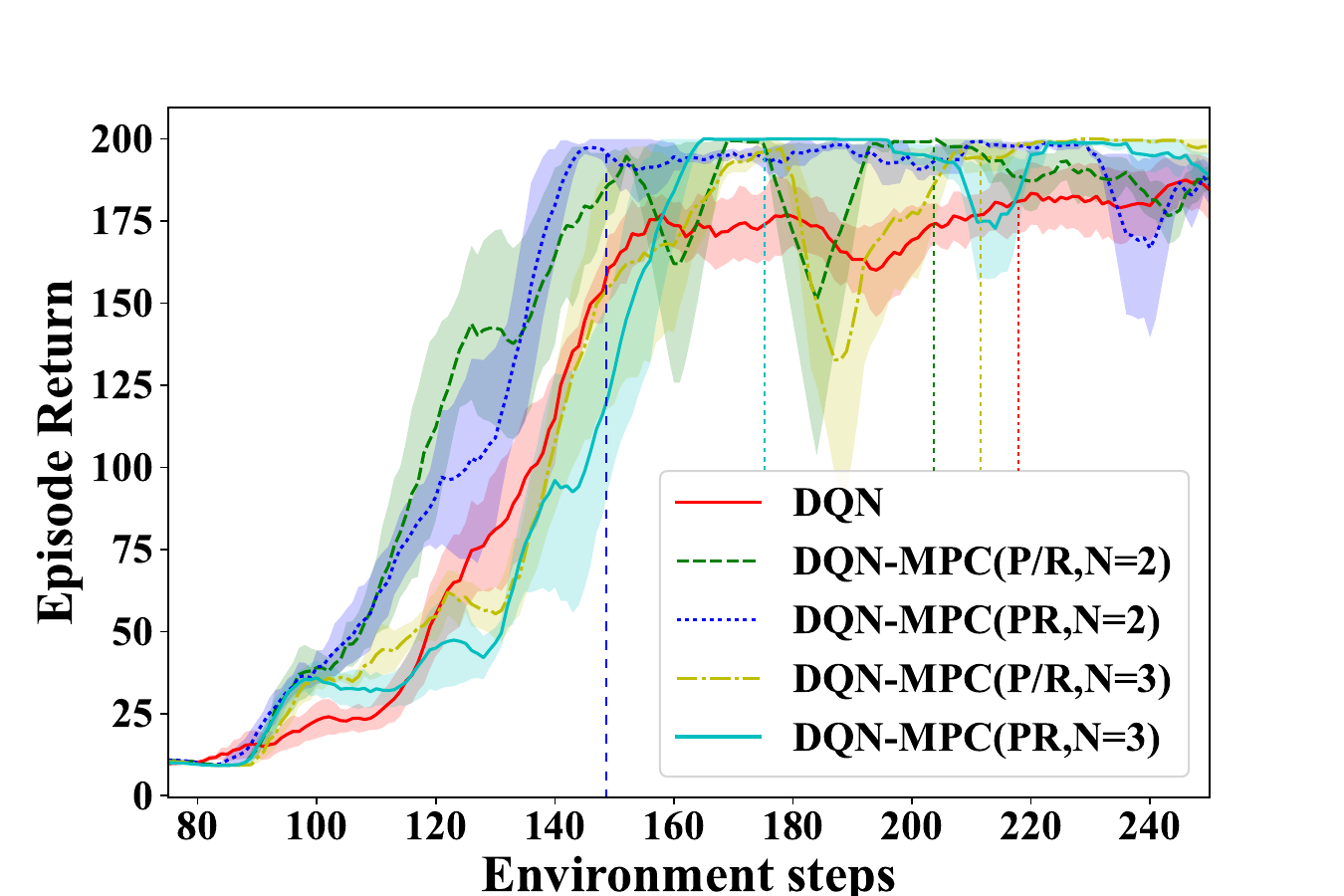} 
}
\subfigure[]{
\includegraphics[width=0.3\textwidth]{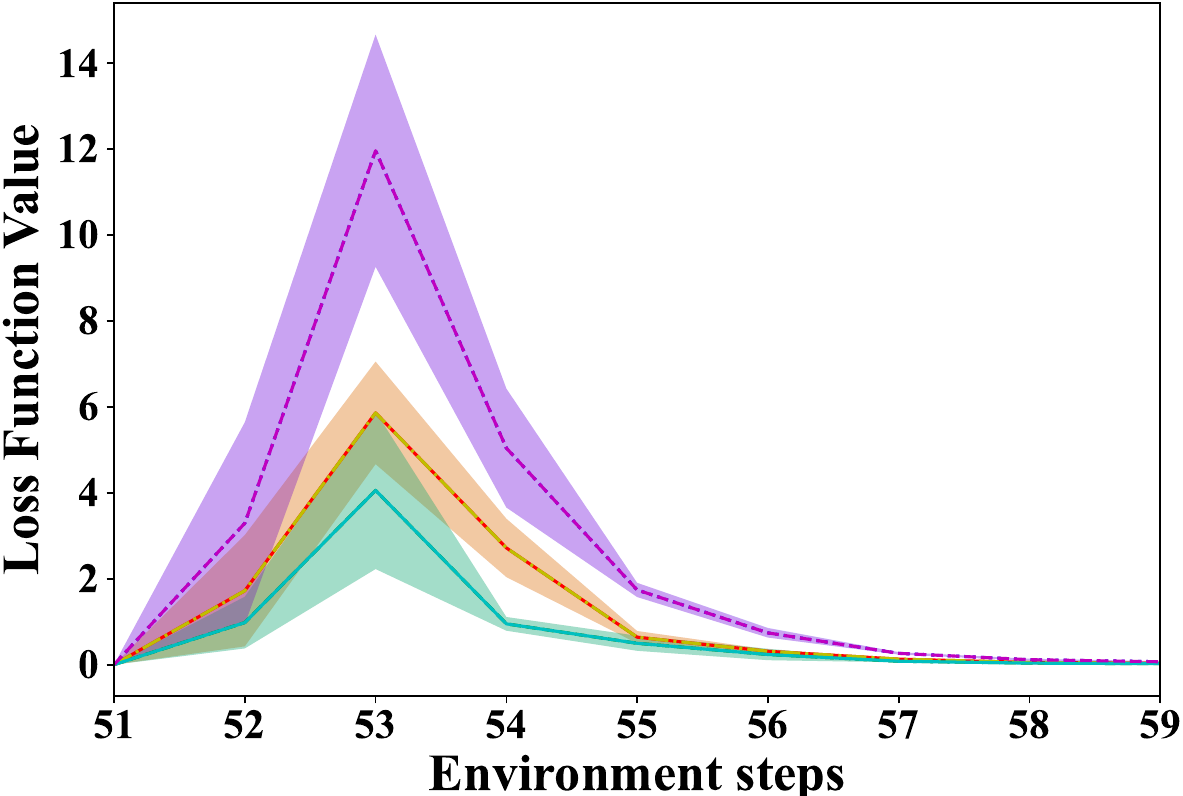}
}
\subfigure[]{
\includegraphics[width=0.3\textwidth]{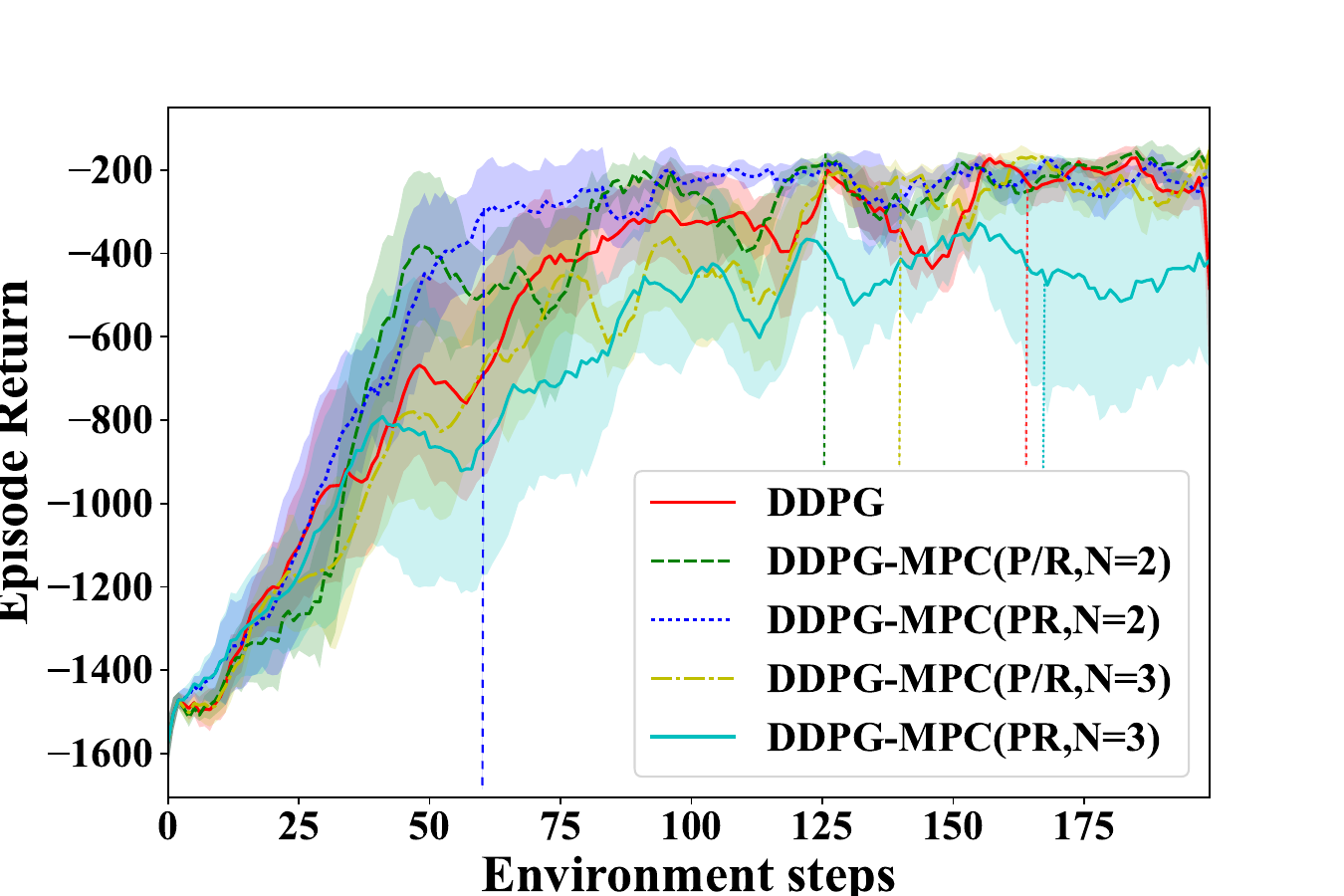}
}
\subfigure[]{
\includegraphics[width=0.3\textwidth]{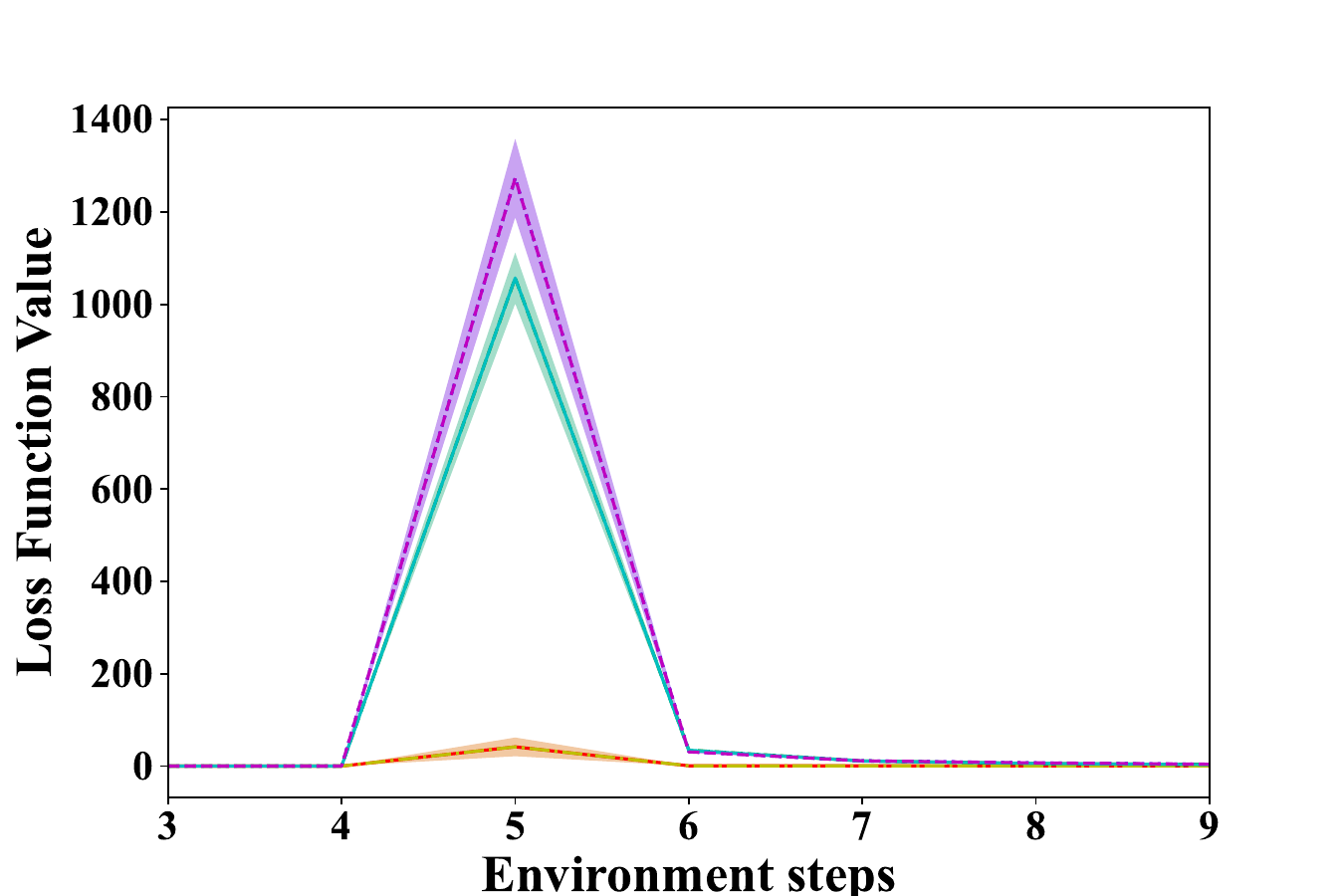} 
}
\subfigure[]{
\includegraphics[width=0.3\textwidth]{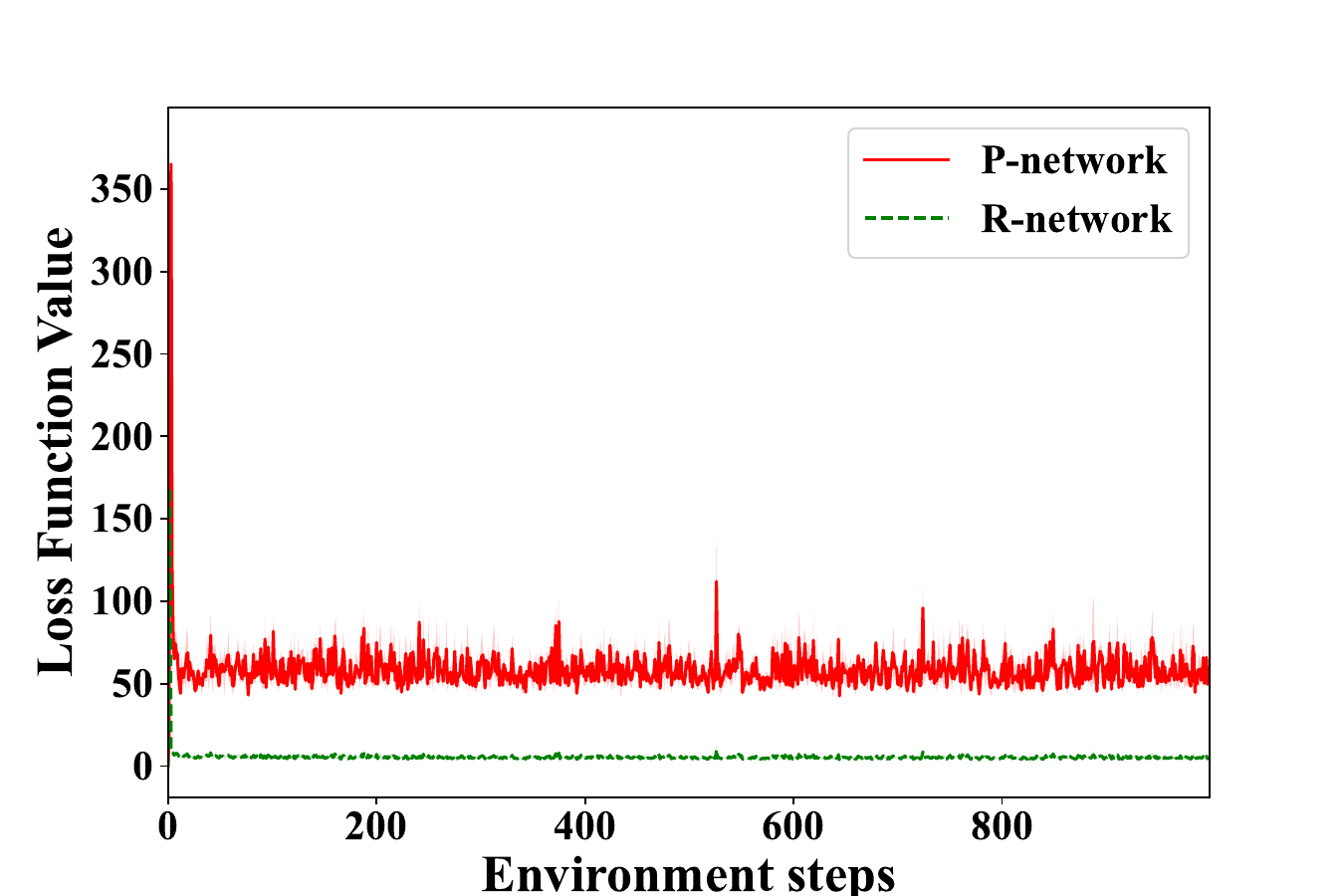} 
}
\subfigure[]{
\includegraphics[width=0.5\textwidth]{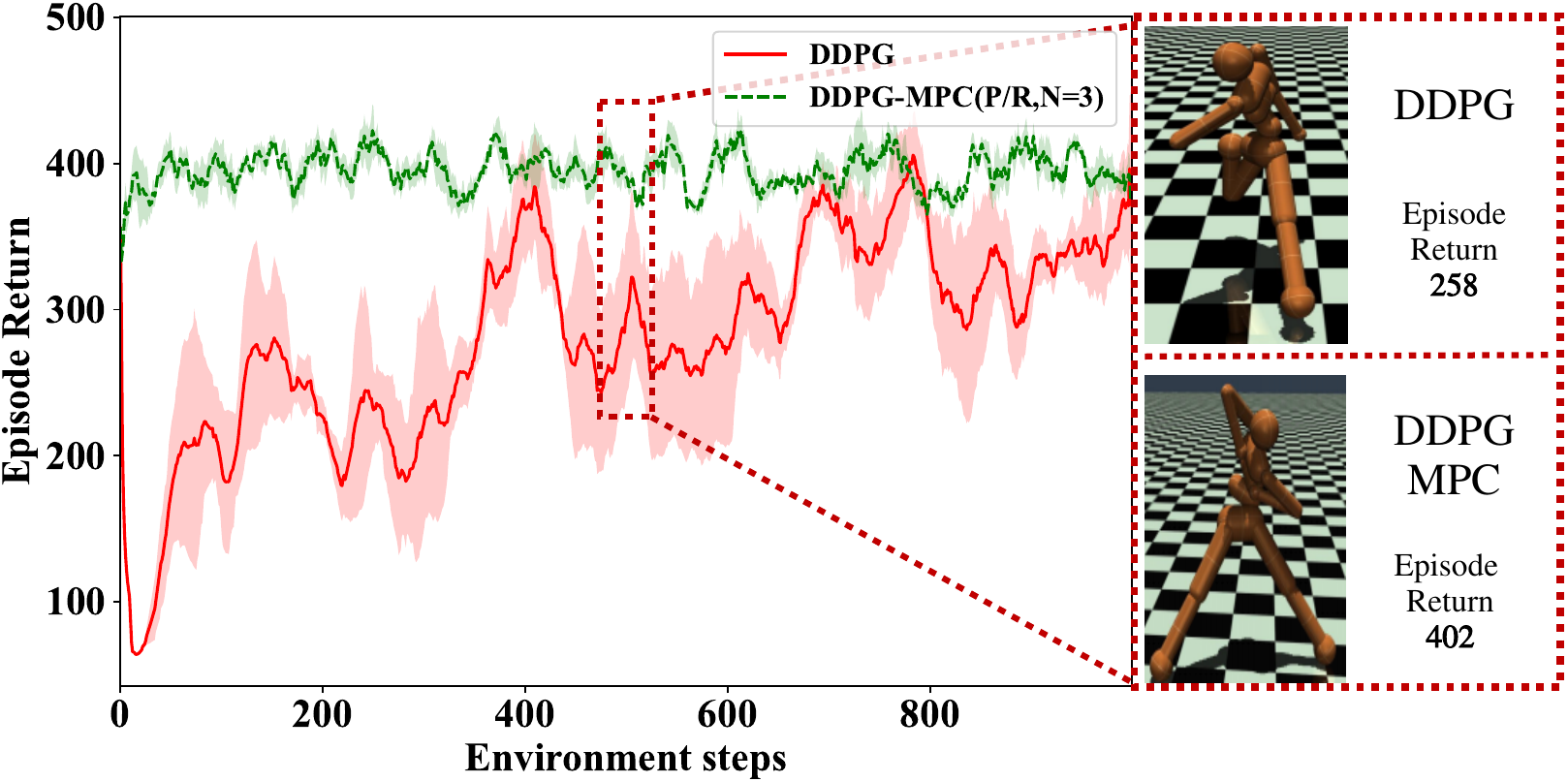} 
}
\DeclareGraphicsExtensions.
\caption{Comparison with MPC--based RL and baseline. (a) Episode Return in CW. (b) Episode Return in CP. (c) Loss Function Value in CP. (d) Episode Return in PD. (e) Loss Function Value in PD. (f) Loss Function Value in HO. (g) Episode Return in HO.}
\label{fig4}
\end{figure*}
Here, we compare our method with traditional RL methods, including $n$--TD$^{4}$, Dyna--q$^{16}$, deep Q--learning (DQN)$^{17}$, and deep deterministic policy gradient (DDPG)$^{18}$, in classic simulation environments and a designed UAV dynamic obstacle avoidance environment. We apply Algorithm \ref{alg2} to obtain Dyna--MPC, DQN--MPC, and DDPG--MPC. Our results demonstrate that our method significantly reduces the samples required for the agent to interact with the environment while achieving local optimal performance.
\subsection{Classic Simulation Environment}
We evaluate our proposed method and compare its performance against baselines on diverse and challenging control tasks from OpenAI Gym. Our experiments are conducted in classic simulation environments, such as Cliff Walking (CW), CartPole (CP), Pendulum (PD), and Humanoid (HO).

\textbf{Implementation details}: We use deterministic components implemented using multi--layer perceptrons. 
The settings for the algorithm parameters are presented in Table \ref{tab1}. We conduct these simulations on a server with a Windows 10 operating system, Intel Core i7--11700 CPU, 16--GB memory, and Radeon 520 GPU. All simulation programs are developed based on Python 3.7 and PyCharm 2022.2.3 compiler. To plot experimental curves, we adopt solid curves to depict the mean of four trials and shaded regions corresponding to standard deviation among trials.

\textbf{Dyna--MPC}: We conduct experiments using $n$--TD and Dyna--q as baselines to implement CW and compare them with Dyna--MPC, as shown in Figure \ref{fig4}(a). For $n$--TD, it is equivalent to Q--learning when $N=1$. As $N$ increases, the curves fall more sharply in the early period but rise to convergence more rapidly. The reason is that $n$--TD adopts multi--step interaction with the environment, and the deviation of the strategy will directly lead to the cumulative deviation of the interaction data over multiple steps. In Dyna--q, it stores $P\left( s,a \right)$ and $R\left( s,a \right)$ in a table similar to $Q\left( s,a \right)$, which provides an accurate model of the environment when the environment is a deterministic process. Therefore, by the model--based approach, its convergence will be faster than Q--learning. Different from $n$--TD, Dyna--MPC performs multi--step prediction in the value estimation relying on deterministic environment models $P\left( s,a \right)$ and $R\left( s,a \right)$. The method avoids the cumulative deviation of interaction data and improves the training efficiency. Through comparisons, Dyna--MPC leads the strategy to quickly converge to the local optimal value based on fewer interaction data. As the prediction step $N$ increases, fewer episodes are required for the strategy to converge to the local optimal value. However, $N=6$ provides limited performance improvement over $N=2$ and requires more computational efforts for multi--step prediction.

\textbf{DQN--MPC and DDPG--MPC}: DQN is adopted as the baseline to implement CP, and the performance is compared with DQN--MPC as shown in Figure \ref{fig4}(b)--(c). $P/R$ or $PR$ denotes modeling $P$ and $R$ as separate neural networks or combining them into one network. The buffer size of DQN--MPC is half of DQN, but it can reduce the number of sample interactions required for the strategy to converge to the local optimal value. In CP ($N^*=2$), a better result is achieved by modeling $P$ and $R$ as one neural network. An additional prediction step $N$ cannot further reduce the number of sample interactions. 

We use DDPG as the baseline to implement PD and HO and compare the performance with DDPG--MPC, as shown in Figure \ref{fig4}(d)--(e). In PD ($N^*=2$), the buffer size of DDPG--MPC is half of DDPG, but it can reduce the number of sample interactions required for the strategy to converge to the local optimal value. Better results are achieved by modeling $P$ and $R$ as one neural network. In HO ($N^*=3$), the buffer size of DDPG--MPC is one--tenth of DDPG. However, it can significantly reduce the sample interactions required for the strategy to converge to the sub--optimal value, which is not the global optimality. The right part of Figure \ref{fig4}(g) shows the average returns from $480^{th}$ episode to $520^{th}$ episode. Since the Mujoco environment has high--dimensional state and action spaces, it is difficult to approximate its state transition and reward functions. 
The learning speed and accuracy of the environment model limit the speed and optimality with which the policy finds the local optimal value.
Through comparisons, DQN--MPC and DDPG--MPC lead the strategy to quickly converge to the local optimal value based on fewer interaction data and free up more sample capacity space required for the experience replay buffer.

\textbf{Model learning of environment}: 
MPC--based RL method learns the state transition and reward functions of the environment while optimizing the policy, as shown in Figure \ref{fig4}. In classical control environments with low--dimension, the loss function values of $P$ and $R$ converge quickly to $0$, indicating that the networks fit the environment well. However, in HO, due to its high dimensionality, the loss function values of the networks cannot converge to $0$. As a result, the networks cannot accurately represent the state transition and reward of the real environment. The deviation of the learned model from the actual environment accumulates in MBRL, leading to only a sub--optimal strategy.
\begin{figure}[htb]
    \centering
    \includegraphics[width=0.4\textwidth]{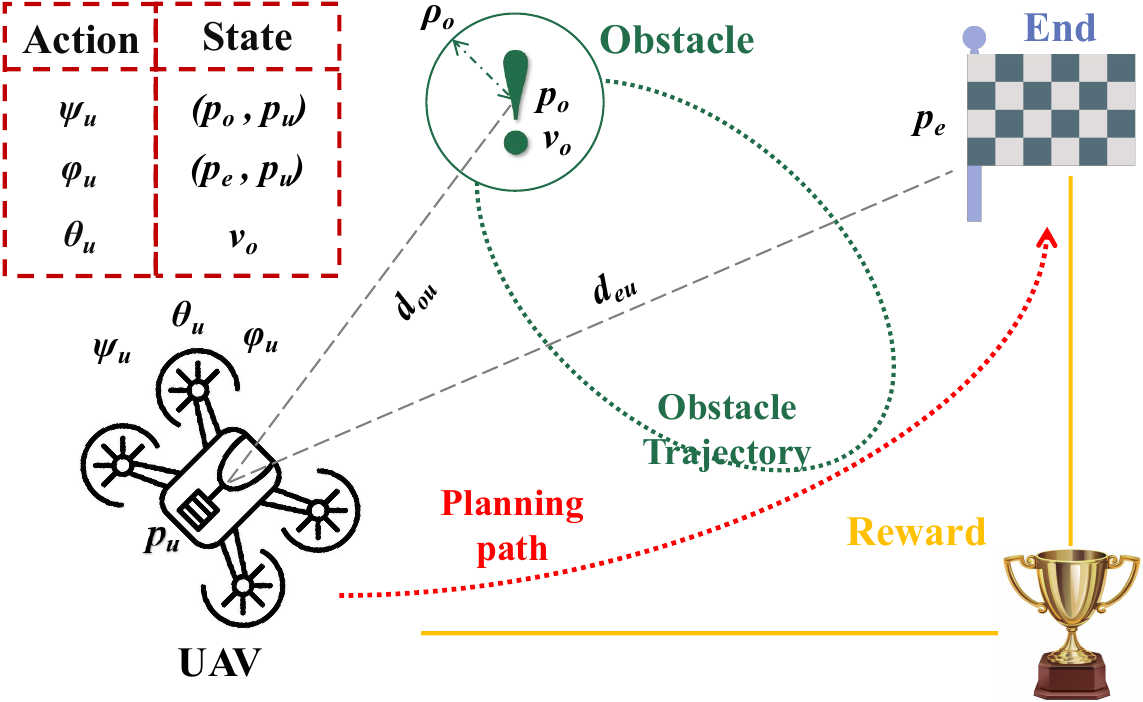}
    \caption{The MDP of UAV dynamic obstacle avoidance problem.}
    \label{fig5}
    \vspace{-1.2em}
\end{figure} 
\subsection{UAV Dynamic Obstacle Avoidance Environment}
To validate MPC--based RL in the path--planning task, we establish the UAV dynamic obstacle avoidance problem as a complete MDP shown in Figure \ref{fig5} and we use the orb as a dynamic obstacle in the scene. 
We simulate the obstacle avoidance in real scenarios through the simulation, including the dynamics of the UAV and the movement of the obstacle.

Let position and velocity vectors in 3D be denoted by $p$ and $v$, respectively. $p_u,p_o,p_e$ denote the positions of the UAV center, obstacle center, and destination, respectively. $v_o$ and $\rho_o$ denote the obstacle's velocity and radius, respectively. $\rho_u$ denotes the radius of the UAV. The state space can be set in the following form:
\begin{align}
        s=\left[ \begin{array}{c}
	\displaystyle\left( p_o-p_u \right) \cdot \frac{d_{ou}-\left(\rho _o + \rho _u\right)}{d_{ou}}\\
	\displaystyle p_e-p_u\\
	\displaystyle v_o\\
\end{array} \right], 
        \label{equ18}
\end{align}
and $d_{ou}=\lVert p_o-p_u \rVert$ is the distance between the UAV and obstacle center, where $\lVert \cdot \rVert$ denotes the Euclidean norm. The action space can be set in the following form:
\begin{figure*}[htb]
\centering
\subfigure[]{
\includegraphics[width=0.38\textwidth]{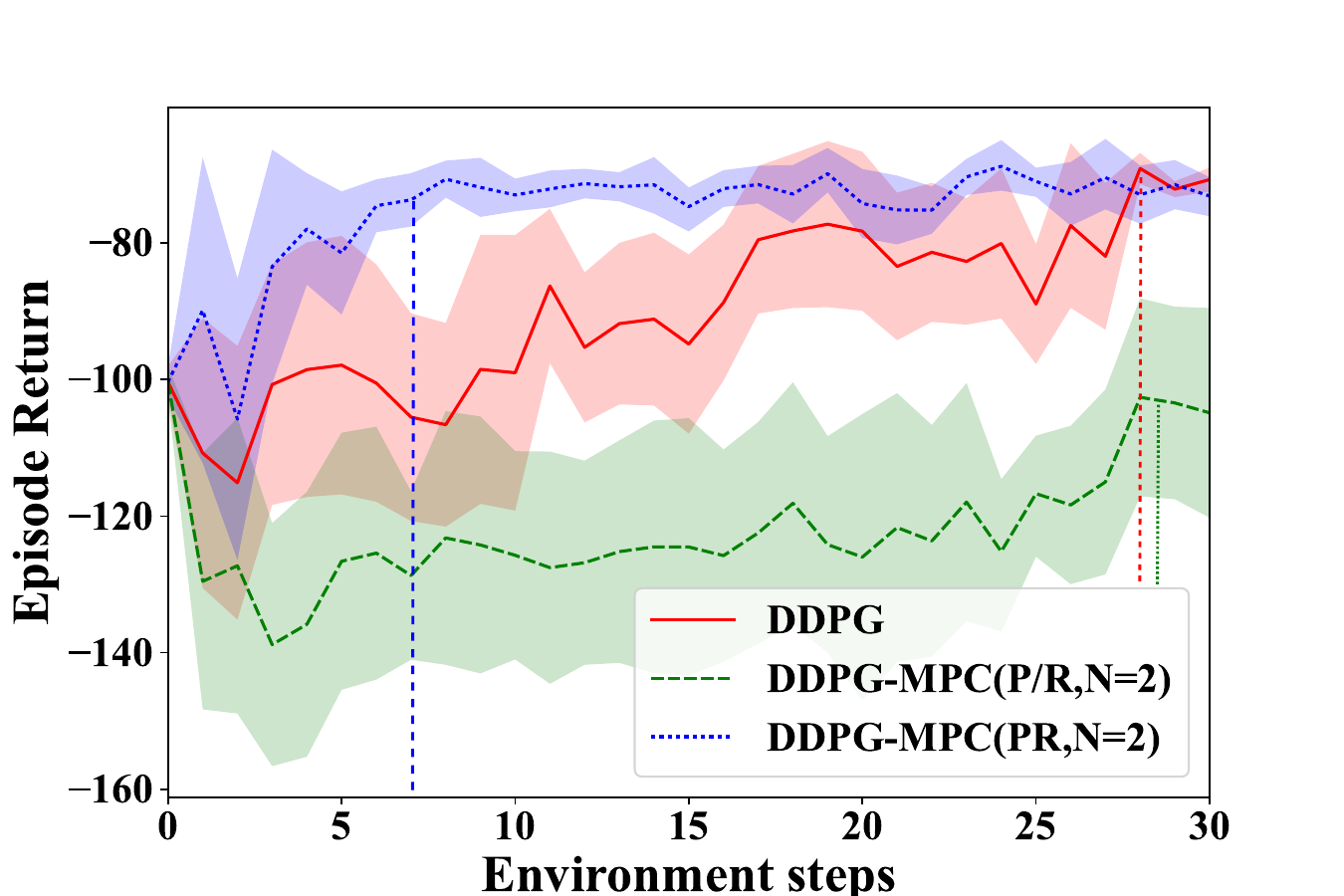} 
}
\subfigure[]{
\includegraphics[width=0.38\textwidth]{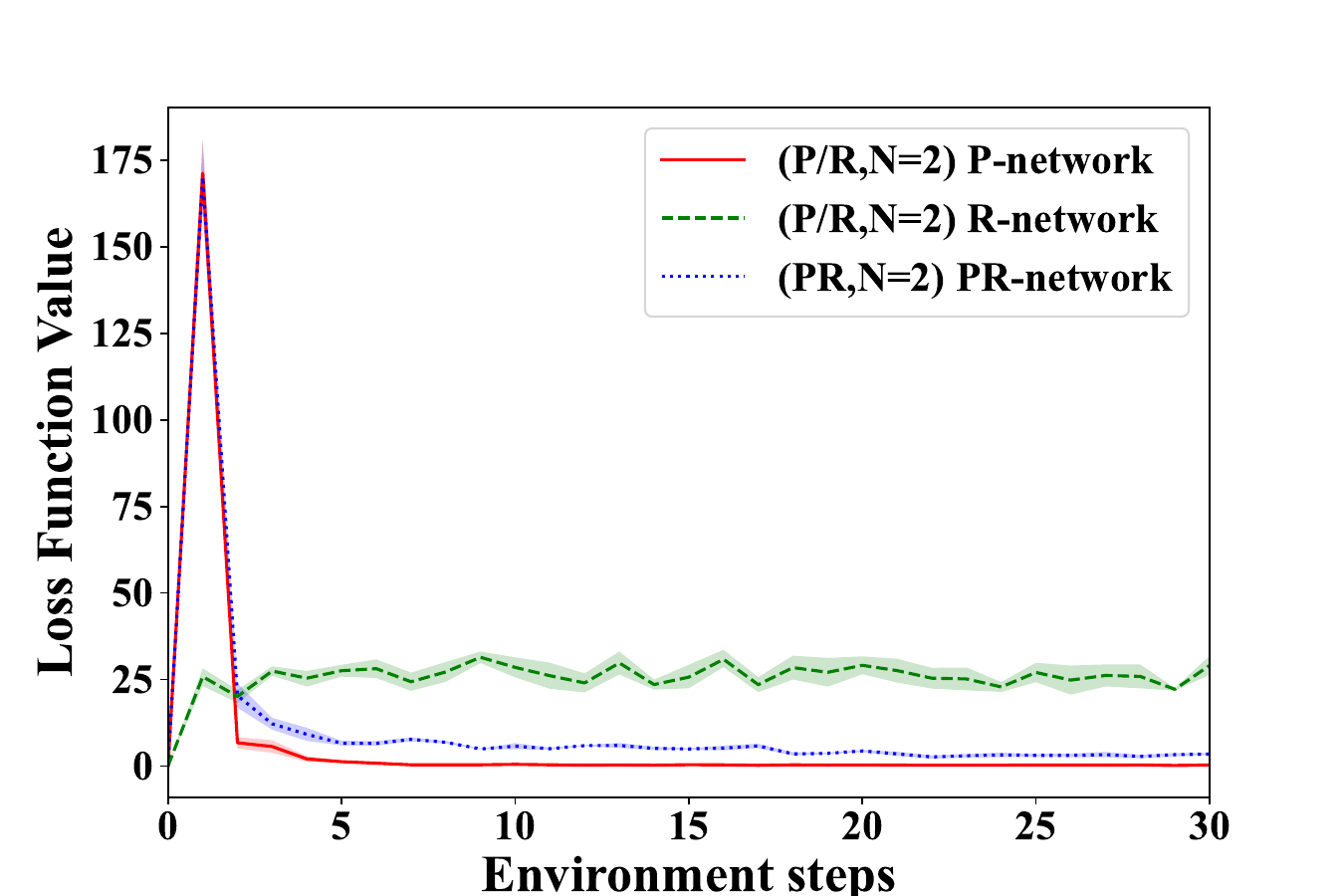} 
}
\subfigure[]{
\includegraphics[width=0.22\textwidth]{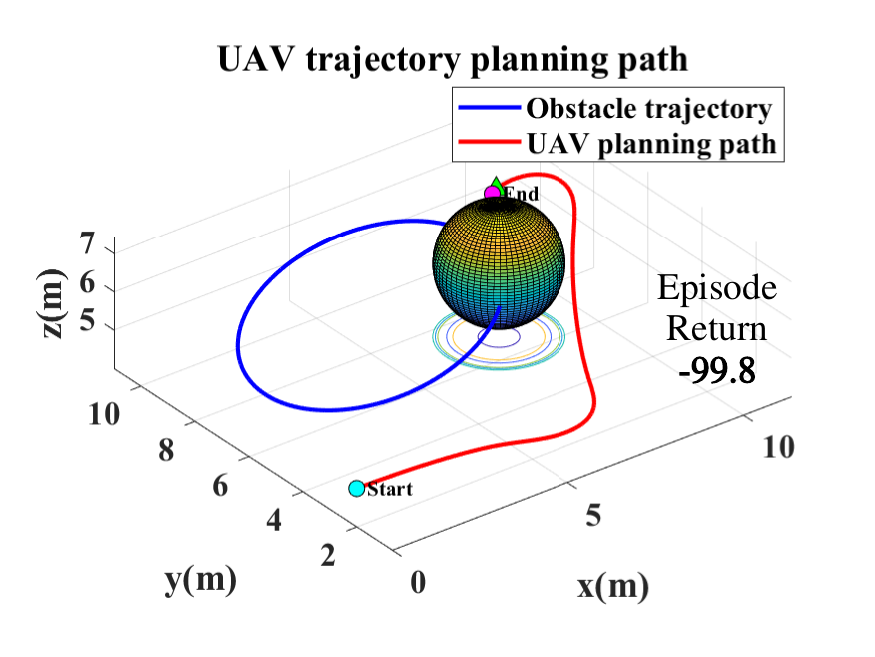} 
}
\subfigure[]{
\includegraphics[width=0.22\textwidth]{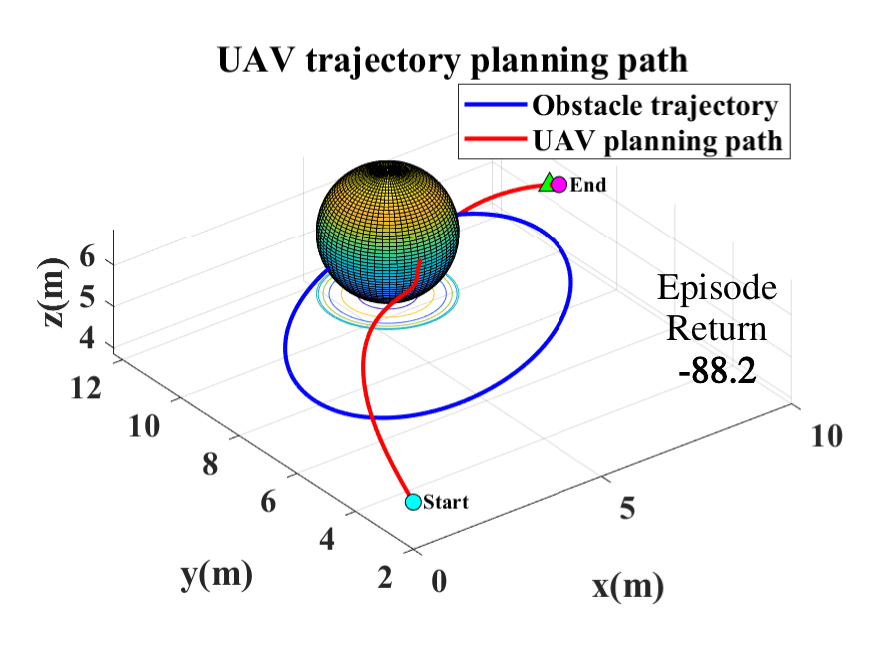} 
}
\subfigure[]{
\includegraphics[width=0.22\textwidth]{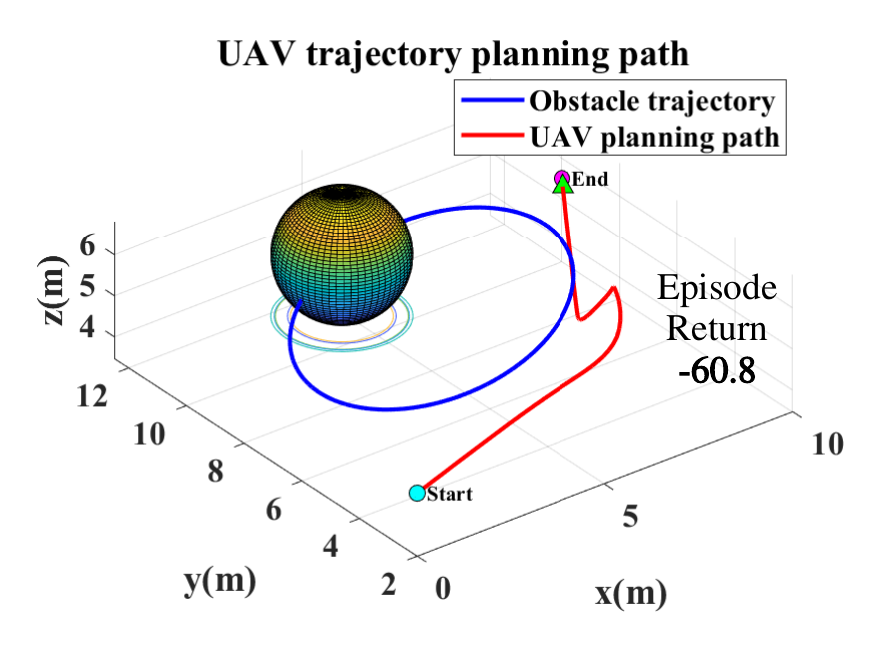} 
}
\subfigure[]{
\includegraphics[width=0.22\textwidth]{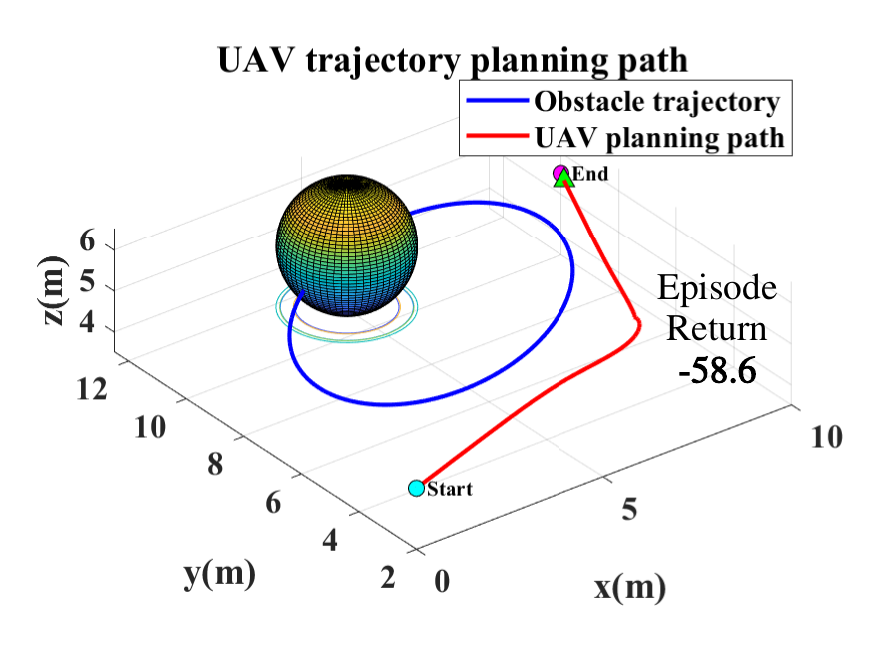} 
}
\subfigure[]{
\includegraphics[width=0.9\textwidth]{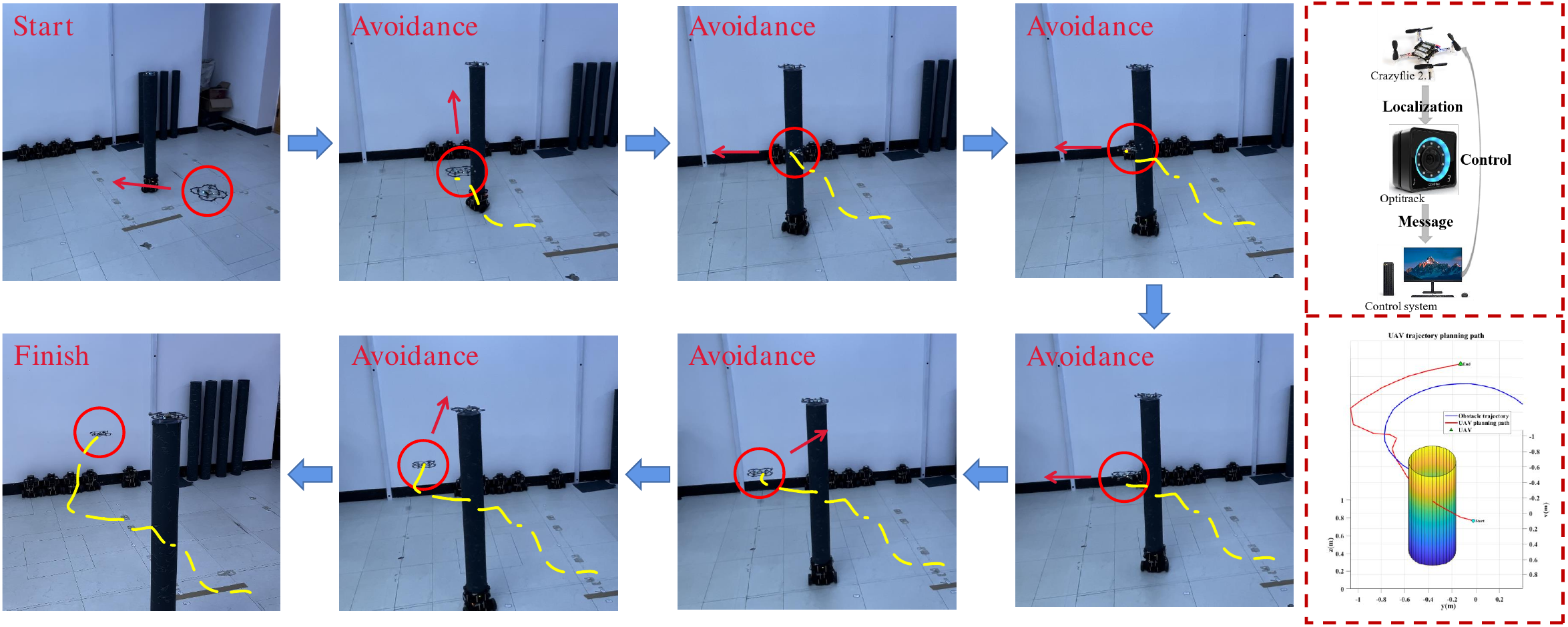}
}
\DeclareGraphicsExtensions.
\caption{Comparison with DDPG in UAV path--planning. (a) Episode Return. (b) Loss Function Value. (c) DDPG $7^{th}$ episode. (d) DDPG $23^{rd}$ episode. (e) DDPG--MPC $7^{th}$ episode. (f) DDPG--MPC $23^{rd}$ episode. (g) Indoor flight verification.}
\label{fig6}
\end{figure*}
\begin{align}
        a=\left[ \begin{matrix}
	\psi _u&		\theta _u&		\varphi _u\\
\end{matrix} \right] ^T,
        \label{equ19}
\end{align}
where $\psi _u,\theta _u,\varphi _u$ denote the UAV's roll, yaw, and pitch, respectively. To lead the UAV to avoid a dynamic obstacle and achieve the shortest path planning, we design the reward as follows:
\begin{align}
        &r=\nonumber \\
	&\begin{cases} \displaystyle\frac{d_{ou}-\left(\rho _o + \rho _u\right)}{\rho _o + \rho _u}-r_a,&		if\,\,d_{ou}<\rho _o + \rho _u\\
	\displaystyle-\frac{d_{eu}}{d_{es}}+r_b+r_c,&		if\,\,d_{ou}>\rho _o + \rho _u\,\,\&\,\,d_{eu}<d_{com}\\
	\displaystyle-\frac{d_{eu}}{d_{es}}+r_c,&		otherwise\\
\end{cases},
        \label{equ20}
\end{align}
where $d_{eu}=\lVert p_e-p_u \rVert$ denotes the distance between the destination and UAV center. $d_{es}=\lVert p_e-p_s \rVert$ denotes the distance between the destination and beginning. $d_{com}$ is an arrival distance, and if $d_{eu}<d_{com}$, the task is completed. $r_a$ is a constant reward and $r_c$ is the additional reward for completing the task. $r_b$ is a threat reward set to keep the UAV as far away from the obstacle as possible:
\begin{align}
        &r_b=\nonumber \\
        &\left\{ \begin{matrix}
	\displaystyle\frac{d_{ou}-\left( \rho _o + \rho _u+d_{thr} \right)}{\rho _o + \rho _u+d_{thr}}-r_d,&		if\,\,d_{ou}<\rho _o + \rho _u+d_{thr}\\
	\displaystyle 0,&		otherwise\\
\end{matrix} \right. ,
        \label{equ21}
\end{align}
where $r_d$ is a constant reward we set. $d_{thr}$ is a threat distance, and if $d_{ou}<\rho _o + \rho _u+d_{thr}$, we consider that the UAV is entering the area where it is about to collide with the obstacle. With the design of the above reward function, when the path is planned with higher cumulative rewards, it can indicate that the path takes less time and has a smaller probability of collision with the obstacle. We apply (\ref{equ18})--(\ref{equ20}) as the state, action, and reward spaces in the MDP, respectively. The parameters are set as follows: $\rho_o=1.5~m$, $d_{thr}=0.4~m$, $r_a=1$, $r_b=3$, $r_d=0.3$. In addition, the state transition of the environment $P\left( s,a \right)$ is implemented based on the disturbed flow field
algorithm and kinematic constraint functions$^{19}$. Next, we can verify the performance of MPC--based RL based on this MDP model.

We adopt DDPG as the baseline to implement UAV path planning and compare its performance with DDPG--MPC ($N^*=2$), as shown in Figure \ref{fig6}(a)--(b). The buffer size of DDPG--MPC is one--tenth of DDPG, but it can significantly reduce the sample interaction required for the strategy to converge to the local optimal value. By modeling $P$ and $R$ as one neural network, the loss function values of the network can converge quickly to 0, leading to better results.
In Figure \ref{fig6}(c)--(f), we compare the path--planning performance of DDPG--MPC and DDPG under different episodes. DDPG--MPC achieves higher cumulative rewards for the same episode than DDPG. Since higher cumulative rewards mean shorter path spending time with less probability of obstacle collisions, DDPG--MPC plans better paths than DDPG. 

\subsection{Indoor Flight Experiment}
We conduct experiments with a real UAV to verify the feasibility of the designed DDPG--MPC approach. In the experiment, we adopt the method of simulation to reality$^{20}$. In an indoor environment, as shown in Figure \ref{fig6}(g), 
we maneuver the crazyflie through the ground control center to complete experiments with motion capture from the optitrack.
We use a plastic tube with a radius of 0.1 meter and a height of 1 meter as a cylindrical obstacle, and UAV traverse the obstacle from one end to the other. By performing DDPG--MPC training in the simulation, we can get the policy network of the UAV in the dynamic obstacle avoidance scenario. Based on the policy network above, the UAV can avoid the obstacle well in the experiment and reach the target position. The experiment's success verifies that DDPG--MPC can accomplish the UAV dynamic obstacle avoidance task by offline training and online decision making, and this RL method is improved based on MPC.
\vspace*{-5pt}

\section{CONCLUSIONS}
\label{sec:sample4}
This paper has presented a novel MPC--based value estimation approach, which improves the training efficiency and sample utilization of the agent in RL. The proposed method has been effective in classical simulation environments and UAV path--planning environments. The method has successfully approximated the model of the environment, which has been well adapted to low--dimensional space. We plan to investigate applying probabilistic ensemble models to learn the environment in high--dimensional space.
\vspace*{-5pt}

\section{ACKNOWLEDGMENTS}
This work was supported in part by the National Key R\&D Program of China under Grant 2022YFC3300703, and in part by the National Science Foundation of China under Grants 62088101 and 62003015. Lei Chen is the corresponding author.
\vspace*{-5pt}

\def\refname{REFERENCES}

\vspace*{-8pt}

\begin{IEEEbiography}{QIZHEN WU}{\,} is pursuing the master degree with the School of Automation Science and Electrical Engineering, Beihang University, 100191, Beijing, China. His research interests include reinforcement learning, robotic control, and task scheduling. He received the B.S. degree in the School of Aeronautics and Astronautics, Sun Yat--sen University, Guangzhou, China. Contact him at wuqzh7@buaa.edu.cn.\vspace*{8pt}
\end{IEEEbiography}

\begin{IEEEbiography}{LEI CHEN}{\,} is currently an associate
research fellow with the Advanced Research Institute of Multidisciplinary Science, Beijing Institute of Technology, Beijing, China. His research interests include complex networks, characteristic model, spacecraft control, and network control. He received the Ph.D. degree in control theory and engineering from Southeast University, Nanjing, China, in 2018. He was a Visiting Ph.D. Student with the Royal
Melbourne Institute of Technology University, Melbourne, VIC, Australia, and Okayama Prefectural University, Soja, Japan. From 2018 to 2020, he was a Post--Doctoral Fellow with the School of Automation Science and Electrical Engineering, Beihang University, Beijing, China. Contact him at bit$\_$chen@bit.edu.cn.\vspace*{8pt}
\end{IEEEbiography}

\begin{IEEEbiography}{KEXIN LIU} {\,}  is currently an Associated Professor with the School of Automation Science and Electrical Engineering, Beihang University, Beijing, China. His research interests include multiagent systems and
complex networks. He received the M.Sc. degree in control and
engineering from Shandong University, Jinan, China, in
2013, and the Ph.D. degree in system theory from the
Academy of Mathematics and Systems Science, Chinese
Academy of Science, Beijing, China, in 2016. From 2016
to 2018, he was a Postdoctoral Fellow with Peking
University, Beijing. Contact him at skxliu@163.com.
\end{IEEEbiography}

\end{document}